# Use of Fuzzy Sets in Semantic Nets for Providing On-Line Assistance to User of Technological Systems

M-N. OMRI* and M.A. MAHJOUB**

*Preparatory institute of engineering studies of Monastir*
*Kairouan Road, 5019 Monastir, Tunisia*
*Nazih.Omri@ipeim.rnu.tn
**MedAli.Mahjoub@ipeim.rnu.tn

**Abstract.** The main objective of this paper is to develop a new semantic Network structure, based on the fuzzy sets theory, used in Artificial Intelligent system in order to provide effective on-line assistance to users of new technological systems. This Semantic Networks is used to describe the knowledge of an "ideal" expert while fuzzy sets are used both to describe the approximate and uncertain knowledge of novice users who intervene to match fuzzy labels of a query with categories from an "ideal" expert. The technical system we consider is a word processor software, with Objects such as "Word" and Goals such as "Cut" or "Copy". We suggest to consider the set of the system's Goals as a set of linguistic variables to which corresponds a set of possible linguistic values based on the fuzzy set. We consider, therefore, a set of interpretation's levels for these possible values to which corresponds a set of membership functions. We also propose a method to measure the similarity degree between different fuzzy linguistic variables for the partition of the semantic network in class of similar objects to make easy the diagnosis of the user's fuzzy queries.

## 1. Introduction

Learning how to use new technological system is mainly an exploratory activity[2]. Exploring learning has shown to improve the abilities constructing to successful error handling and discovering, and eventually constructing, correct knowledge. But exploratory activity frequently leads to experience uninterested states does not reach the interested state goal. Users need assistance not only to avoid errors, but also to understand how the system interprets their commands [8, 9, 12].
In order to respond to a query, an executive assistant might know very precisely the goal the user has in mind [14], which means an object in a given state (the properties of the object being transformed). Moreover, even when goals are fairly well defined, it is often necessary to think about superordinate goals.
The fuzzy set method has been used to develop the "on-line instructions" mechanisms of an Intelligent Assistance System. It can be seen as a supervisor of task execution that has the "ideal user's knowledge" of (i) prerequisites of procedures, (ii) subGoals structure[3]. And (iii) the semantic network of the elements of the device where applied procedures are used as properties, as well as (iv) the knowledge of perceptible and imperceptible effects of user's actions[4]. With an interactive dialogue with a user, the Assistance System tries to match items provided by users in natural language with the knowledge included in the ideal user's semantic network [3].

## 2. Expert Semantic Net

The example of the technical system we consider here is Word Processor software (figure1), with Objects such as "*chain-of-characters*", and procedures such as "*cut*" or "*copy*". For a novice user of the software, the list of standard denominations is not obvious and he often would like to ask an expert operator about how to execute an action such as "*how to rub letters*" [4].
The underlying psychological hypothesis is that Goals are Object properties, and are generators of Object categories. Goals and procedures define the function of Objects and the way to use them. As functional properties of Objects, they enter into the construction of semantic networks in the same way as structural properties. We define a procedure as a sequence of operations whose execution serves to reach a Goal, and where the elements of the sequence are either primitive actions or subGoals which themselves call for associated procedures.

## 3. Linguistic variables and membership function

In mathematical treatment of the linguistic variable aiming to process some fuzzy deductions by computers, a linguistic variable definition imposes itself. In this context, we assign to every value of the linguistic variable a membership functions $\mu$, the value varies between 0 and 1. While holding amount of the classification in a certain number of fuzzy sets, this represents the fuzzyfication process. For the representation of the different concepts of our system, we propose to consider the set of system's Goals (respectively Objects) as a set of linguistic variables.

### 3.1. The levels of interpretation and its membership functions

So far, we have introduced a fine subdivision, with four or five values for the linguistic variable that we manipulate forming thus as many fuzzy sets as it is shown in Fig. 1. We have also defined five levels of interpretation because actually, it appears that, for more than five levels, we have a problem of natural discrimination between the different levels and less than five is insufficient to have a good discrimination. Our objective is to have a cleaner idea of the interpretation of a user's Goal with regard to the fuzzy knowledge basis.

| Levels of interpretation | Membership function |
|---|---|
| *It's Not True* | [0,0,0.2,0.4] |
| *It's Little True* | [0.2,0.4,0.4,0.6] |
| *It's Half True* | [0.4,0.6,0.6,0.8] |
| *It's Rather True* | [0.6,0.8,0.8,1] |
| *It's Quite True* | [0.8,1,1,infinity] |

Table 1. Different levels of interpretation and their corresponding membership function.

We can distinguish, therefore, five possible values for a given linguistic variable and therefore five membership functions corresponding to the sets of values in table 1. These values are determined to the departure by the expert, and adjusted progressively by the system when a user's request has been identified with success.

### 3.2. Definition of the fuzzy linguistic variables

Let's consider the System's Goal *CutWithKey* as a linguistic variable for a novice user of the system. We can distinguish five values "*not_true*", "*little_true*", "*half_true*", "*rather_true*" and "*quite_true*" to which corresponds five fuzzy sets (Fig. 1).
So *CutWithKey* of 0.25 belongs with a membership factor $\mu$=0.7 to the "*half_true*" set and with $\mu$=0.3 to the "*quite_true*" set. Explicitly, we can write $\mu_{half\_true}$(*CutWithKey*=0.25)=0.7 and $\mu_{quite\_true}$(*CutWithKey*=0.25)=0.3.
We consider the membership function of the verb *to Gum* with regard to the Expert Goal *CutWithKey*. The Fig. 1 and Fig. 2 represent the result of the adjustement process. So *CutWithKey* of 0.75 belongs with a membership factor $\mu$=0.7 to the "*half_true*" set and with $\mu$=0.3 to the "*quite_true*" set. Explicitly, we can write $\mu_{half\_true}(CutWithKey=0.75)=0.7$ and $\mu_{quite\_true}(CutWithKey=0.75)=0.3$.

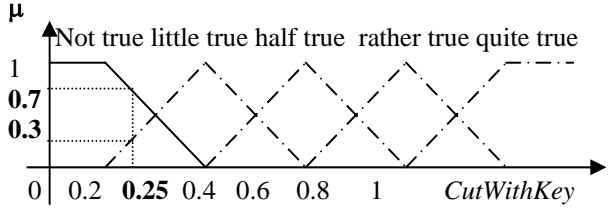
Fig. 1. Membership function of *CutWithKey*

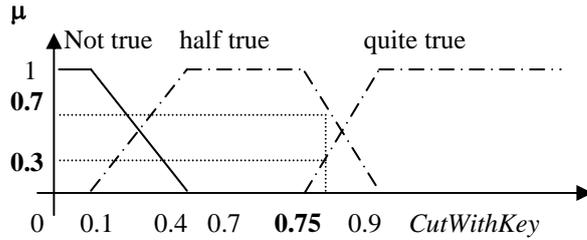
Fig. 2. Membership function of *Gum* with regard to *CutWithKey*

These two membership functions (Fig. 1 and 2) are represented in memory of the computer by the following structure:

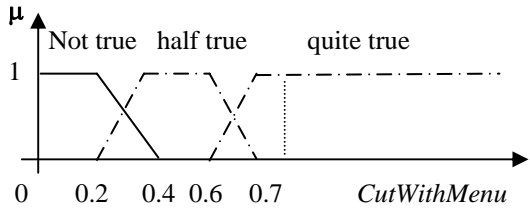
Fig. 3. Membership function of *Gum* with regard to *CutWithMenu*

$Gum_{CutWithMenu}$ = {(not_true, [0,0,0.2,0.4]), (half_true, [0.2,0.4,0.4,0.6]), (quite_true, [0.6,0.7,1,1])}.
$Gum_{CutWithKey}$ = {(not_true, [0,0,0.1,0.4]), (half_true, [0.1,0.4,0.7,0.9]), (quite_true, [0.7,0.9,1,1])}.
On the other hand, $\mu_{quite\_true}(CutWithMenu=0.75)=1$ (Fig. 3).

### 3.3. The structure of user's goal

Let *G* be the user's Goal, we search to identify with regard to the set of a system procedure $P_j$. $f_{ji}$ are the set of membership functions corresponding to the set of different levels of interpretation $L_{ji}$.
The structure of the user's Goal *G* is given by:

$$G = \{(P_j, \{(L_{ji}, f_{ji})/1 \le i \le n\})/1 \le j \le n, n \ge 1\} \quad (1)$$

Where

$L_{ji}$: is the $i^{th}$ level of interpretation relatively to the system's procedures $P_j$,
$f_{ji}$: is the corresponding membership function.

## 4. The attribute's structures of fuzzy concepts

We distinguish between tow types of attributes: system's Goal (respectively Object) attributes and user's Goal (respectively Object) attributes. In this paper we concentrates only on the Goals case.

### 4.1. The structure of the system's goal attribute

Let $P_j$ the set of a system's Goals given by the Expert with a certain equivalence with regard to each system's Goal which can be applied on a given Object $O$, $d_j$ is the set of possibility degree given by the expert. We define the structure of a system's attribute $A^S$ by:

$$A^S = \{(P_l, (p_j, d_j)/1 \leq j \leq n, n \geq 1)/1 \leq l \leq m, m \geq 1\} \quad (2)$$

**Example 2**

$A^S = \{$ (**CutWithMenu**, (CutwithKey, .9), (ErasewithMenu. .6) ),
(**EraseWithKey**, (CutwithKey, .5), (ErasewithMenu.8))
$\}$.

### 4.2. The structure of the user's goal attribute

Let $P_k$ the set of a system's Goals be identified with success to a given user's Goal $G_l$, $f_{ji}$ is the set of membership functions corresponding to the set of different levels of interpretation $L_{ji}$. We define a user's attribute $A^U$ by:

$$A^U = \{(G_l, (p_k, \{(L_{ki}, f_{ki})/1 \leq i \leq n\})/1 \leq k \leq n)/1 \leq l \leq m, n, m \geq 1\} \quad (3)$$

**Example 3**

$A^U = \{$ (**to_Gum**, (**CutwithKey**, (not_true, [0,0.1,0.3,0.4]), (half_true, [0.3,0.6,0.7,0.8]), (quite_true, [0.7,0.8,0.8,1])), (**ErasewithMenu**, (not_true , [0,0,0.1,0.4]), (half_true, [0.2,0.4,0.4,0.6] ), (quite_true, [0.4,0.9,0.9,1]))),
(**to_Rub**, (**CutwithKey**, (half_true, [0,0.2,0.3,0.5]), (rather_true, [0.3,0.5,0.6,0.8 ]), (quite_true, [0.6,0.8,0.9,1] ) ), (**ErasewithMenu**, (not_true, [0,0,0.2,0.4]), (half_true, [0.2,0.4,0.4,0.6] ), (rather_true, [0.7,0.9,0.9,1])))
$\}$.

The integration of fuzzy properties; as "*to Gum*", in the Object's description, implies the valuation of relations in [0, 1] [3, 4]. There are two kinds of relationships: the relation "*kind-of*" between two classes and the relation "*is-a*" between a class and an instance. One class may be a kind-of an other class, to some extend. Each kind of relationship is described by a membership value obtained from the inclusion between areas or between attributes.

## 5. The hierarchical fuzzy relations

In the case where the X universe is discrete, the degree of inclusion is given by [Zadeh1978]. If we consider that A and B are two fuzzy subsets of X, the inclusion degree of A in B is given by :

$$Deg(A \subset B) = \frac{\sum_{x \in X} f_{A \cap B}(x)}{\sum_{x \in X} f_A(x)} \quad (4)$$

### 5.1. Case of the fuzzy system's linguistic variables

Let a and b be two system's linguistic variable defined on the same universe of procedures P by $a=\{(a_1,d_1),(a_2,d_2),...,(a_n,d_n)\}$ and $b=\{(b_1,d_1'),(b_2,d_2'),...(b_n,d_n')\}$. Then the inclusion's degree of a in b is given by :

$$Deg\ (a \subset b) = \frac{\sum_{p \in P} f_{a \cap b}(p)}{\sum_{p \in P} f_a(p)} \quad (5)$$

With $f_{a \cap b} = \min_{1 \le i \le n}(d_i, d_i')$ the last formula becomes :

$$Deg\ (a \subset b) = \frac{\sum_{i=1}^{n} \min_{1 \le i \le n}(d_i, d_i')}{\sum_{i=1}^{n} d_i} \quad (6)$$

### 5.2. Case of the fuzzy user's linguistic variables

Let T and S two linguistic variables be defined on the universe of procedures P by :

$$T = \{(P_i, \{(L_{ij}, f_{ij}^T) / 1 \le j \le n\}) / 1 \le i \le n, n \ge 1\}$$
$$\text{and } S = \{(P_i, \{(L_{ij}, f_{ij}^S) / 1 \le j \le n\}) / 1 \le i \le n, n \ge 1\}$$

We have :

$$f_{T \cap S}(p_j) = \min_{1 \le i \le n}(f_i^T \cap f_i^S)(p_j) \text{ and } f_T(p_j) = \min_{1 \le i \le n}(f_i^T)(p_j)$$

Let $f_T$ the membership function associated to the linguistic variable *T*, and $f_{T \cap S}$ be the membership function results from the intersection of *T* and *S*. We define inclusion degree of *T* in *S* by:

$$Deg\ (T \subset S) = \frac{\sum_{p_j \in P} \min_{1 \le i \le n}(f_i^T \cap f_i^S)(p_j)}{\sum_{p_j \in P} \min_{1 \le i \le n}(f_i^T)(p_j)} \quad (7)$$

### 5.3. The inclusion's degree between two fuzzy attributes

Let $T_1, T_2, \ldots, T_i$, i linguistic variables for an attribute A and $S_1, S_2, \ldots, S_i$, i linguistic variables for an attribute B where $T_1 S_1, T_2 S_2, \ldots, T_i S_i$ be defined on the same universe P. The inclusion's degree of A in B is given by:

$$Deg\ (A \subset B) = \frac{1}{k} \sum_{1 \le l \le k} Deg\ (T \subset S)(g_k) \quad (8)$$

### 5.4. The inclusion's degree between two fuzzy classes

Let $A_1, A_2, \ldots, A_n$ be n attributes which defines the fuzzy class $C_1$ and $B_1, B_2, \ldots, B_n$, n attributes which define the fuzzy class $C_2$. We define inclusion degree of $C_1$ in $C_2$ by :

$$Deg(C_1 \subset C_2) = \frac{1}{n} \sum_{1 \le l \le n} Deg(A_i \subset B_i) \quad (9)$$

## 5.5. The membership's degree of an instance to an object's class

In the case of a class and instance, we deal with degrees of membership degrees. These degrees measure the physical representation of the class by the instances. They are obtained from inclusion degrees between fuzzy attributes. We define the membership of an instance I in class C by :

$$Deg\,(I \in C) = \frac{1}{n} \sum_{1 \leq i \leq n} Deg\,(a_i \in A_i) \quad (10)$$

## 5.6. The defuzzification process

Defuzzification is the process of conversion of a fuzzy quantity represented by a membership function to a precise value. In this study, the center of gravity method [10], will be used to defuzzify the outputs membership functions into precise values.

**Example 4 : case of two user's linguistic variables**

**To_Gum:** {
(**CutwithMenu**, (not_true,[0,0,0.1,0.4]), (half_true, [0.2,0.3,0.4,0.6] ), (quite_true, [0.7,0.9,0.9,1]) ),
(**CutwithKey**, (not_true, [0,0,0.1,0.4]), (half_true, [0.2,0.4,0.4,0.6]), (quite_true, [0.7,0.9,0.9,1]))
}.

**To_Rub**: {
(**CutwithMenu**, (not_true, [0,0.1,0.1,0.4] ), (half_true, [0.2,0.4,0.4,0.6]), (quite_true, [0.7,0.8,0.9,1]) ),
(**CutwithKey**, (not_true, [0,0.2,0.3,0.4]), (half_true, [0.2,0.3,0.5,0.6]), (quite_true, [0.6,0.7,0.9,1]))
}.

Firstly, we calculate the gravity center for each membership function, which allows us to calculate the degree of similarity between *To_Gum* and *To_Rub*. We have next results:

**To_Gum** : {
(**CutwithMenu**, (not_true, .14)  (half_true, .38) (quite_true, .86)),
(**CutwithKey**, (not_true, .14), (half_true, .4), (quite_true, .86))

**To_Rub** : {
(**CutwithMenu**, (not_true, .16), (half_true, .4), (quite_true, .85)),
(**CutwithKey**, (not_true, .22), (half_true, .4), (quite_true, .8))
}.

*We then applied the next formula with T and S replaced respectively by To_Gum and To_Rub :*

$$Deg\,(T \subset S) = \frac{\sum_{p_j \in P} \min_{1 \leq i \leq n} (f_i^T \cap f_i^S)(p_j)}{\sum_{p_j \in P} \min_{1 \leq i \leq n} (f_i^T)(p_j)} \quad (11)$$

The system learns by interpreting an unknown word, using the links provided by the context of the query, and created between this new word and known words. With the learning of new words in natural language as the interpretation which was produced in agreement with the user, the system improves its representation scheme at each experiment with a new user and, in addition, takes advantage of previous discussions with users. Then, to make easy the diagnosis of the user's fuzzy queries, we suggest in the next section to deel with the similarity degree [1, 17, 18] between different fuzzy linguistic variables for the partition of the semantic network in class of similar objects.

# 6. The measurement of similarity between fuzzy concepts

The degree of similarity has obviously to be calculated between two Objects with similarly nature, generics or individual[5]. The similarity relationship we have used is given by [20].
The degree of similarity between two fuzziness linguistic concepts is obtained from values of the membership functions associated to their attributes.

## 6.1. The similarity degree between two fuzzy linguistic variables

Let P be the universe of linguistic values, $G$ and $H$ are two User's linguistic variables defined on P such as $G = \{(P_j, \{(L_{ji}, f_{ji}) / 1 \leq i \leq 5\} / 1 \leq j \leq n, n \geq 1\}$ and $H = \{(P_j, \{(L_{ji}, f'_{ji}) / 1 \leq i \leq 5) / 1 \leq j \leq n, n \geq 1\}$. $f_{ji}$ and $f'_{ji}$ are respectively the corresponding membership functions to $L_{ji}$ relatively to system's procedures $P_j$.

Let $f_{G \cap H}$ and $f_{G \cup H}$, respectively the consequent membership function of the intersection, and the union of membership functions associated to $G$ and $H$. We define the degree of similarity between $G$ and $H$ by:

$$Sim(G, H) = \frac{\max_{x \in X} f_{G \cap H}(x)}{\max_{x \in X} f_{G \cup H}(x)} \quad (12)$$

$G$ and $H$ are perfectly similar if and only if: $Sim(G, H) = 1$.

Where $f_{G \cap H}$ and $f_{G \cup H}$ are calculated respectively by formulas 11 and 12 as follows:

## 6.2. The similarity degree between two fuzzy attributes

### 6.2.1. The case of system's attributes

Let $U_j$ and $V_j$, $j \in [1, m]$ the sets of linguistic values of the system's attribute $A^S$ and $B^S$ respectively. $A^S$ and $B^S$ are respectively the properties of the objects $O$ and $O'$. We define the degree of similarity between $A^S$ and $B^S$ by:

$$Sim(A^S, B^S) = \frac{1}{m} \sum_{j=1}^{m} Sim(U_j, V_j) \quad (13)$$

### 6.2.2. The case of user's attributes

Let $U_i$ and $V_i$, $i \in [1, p]$ the sets of linguistic values of the system's attribute $A^U$ and $B^U$ respectively. $A^U$ and $B^U$ are respectively the properties of the objects $O$ and $O'$. We define the degree of similarity between $A^U$ and $B^U$ by:

$$Sim(A^U, B^U) = \frac{1}{p} \sum_{i=1}^{p} Sim(U_i, V_i) \quad (14)$$

The degree of similarity between these two objects is calculated with the similarity degrees between the attributes of these objects.

## 6.3. The similarity degree between two fuzzy objects

We come to define the degree of similarity between two fuzzy attributes $A$ and $B$ of two Objects $O$ and $O'$ respectively. This allows us to obtain the degree of similarity between $O$ and $O'$. It is given by:

Let $O$ and $O'$ two fuzzy Objects such as $O = \{A_1...A_i\}$ and $O' = \{B_1...B_n\}$ where $A_1, B_1...A_i, B_n$ are respectively defined on the same universe X. The degree of similarity between $O$ and $O'$ is given by:

$$Sim(O, O') = \min_{1 \leq k \leq n} sim(A_k, A_k) \quad (15)$$

### 6.4. Example 5

In this example we consider the descriptions of two different user's Goals *to Gum* and *to Rub*, with regard to the Expert's Goals *CutwithMenu* and *CutwithKey* with the following description respectively:

**To_Gum**: {
(**CutwithMenu**, (not_true, [0,0,0.1,0.4]), (half_true, [0.2,0.3,0.4,0.6]), (quite_true, [0.7,0.9,0.9,1])),
(**CutwithKey**, (not_true, [0,0,0.1,0.4]), (half_true, [0.2,0.4,0.4,0.6]), (quite_true, [0.7,0.9,0.9,1]))
}.

**To_Rub:** {
(**CutwithMenu**, ( not_true, [0,0.1,0.1,0.4]), (half_true, [0.2,0.4,0.4,0.6]), (quite_true, [0.7,0.8,0.9,1])),
(**CutwithKey**, (not_true, [0,0.2,0.3,0.4]), (half_true, [0.2,0.3,0.5,0.6]), (quite_true, [0.6,0.7,0.9,1]))
}.

Firslty, we calculate the gravity center for each membership function, which allows us to calculate the degree of similarity between *To_Gum* and *To_Rub*.

**To_Gum** : {(**CutwithMenu**, (not_true, .14) (half_true, .38) (quite_true, .86) ),
(**CutwithKey**, (not_true, .14), (half_true, .4), (quite_true, .86))

**To_Rub** : {(**CutwithMenu**, (not_true, .16), (half_true, .4), (quite_true, .85)),
(**CutwithKey**, (not_true, .22), (half_true, .4), (quite_true, .8))}.

Then the degree of similarity between *to_Gum* and *to_Rub* is calculated as follows:

$$f_{Gum \cap Rub}(CutWithMenu) = \frac{1}{3}(\min(.14,.16) + \min(.38,.4) + \min(.86,.85)) = .46$$

$$f_{Gum \cap Rub}(CutWithKey) = \frac{1}{3}(\min(.14,.22) + \min(.4,.4) + \min(.86,.6)) = .38$$

$$f_{Gum \cup Rub}(CutWithMen) = \frac{1}{3}(\max(.14,.16) + \max(.38,.4) + \max(.86,.85)) = .47$$

$$f_{Gum \cup Rub}(CutWithKey) = \frac{1}{3}(\max(.14,.22) + \max(.4,.4) + \max(.86,.6)) = .49$$

Then

$$Sim(Gum, Rub) = \frac{\max(.46,.38)}{\max(.47,49)} = .94$$

To conclude, we can say that *Gum* and *Rub* are similar by 94 percent.

## 7. Conclusion and future works

In this paper, we have presented a new structure of semantic network based on fuzzy sets theory. The System has been tested on available databases. We have compared our system with neural network-based approaches and with other semantic net-based techniques. Experimental results, that will be published later, have shown the effectiveness of the approach proposed in providing effective on-line assistance to users of new technological systems. This approach can serve as a basis for our research to elaborate a general methodology to diagnosis the purpose Goal of the subject, applicable to a large diversity of Objects which allow a best approximation of the category of the purpose aimed by the user and best approaches the diagnosis.